\documentclass[runningheads]{llncs}
\usepackage[T1]{fontenc}
\usepackage{graphicx}
\usepackage[table]{xcolor} 

\usepackage{adjustbox}
\usepackage{pgfplots}
\usepackage{tikz}
\usetikzlibrary{calc}
\usepackage{subcaption}
\usepackage{booktabs}
\usepackage{multirow}
\usepackage{multicol}
\usepackage{adjustbox}
\pgfplotsset{compat=1.18}
\usepgfplotslibrary{statistics}
\usepackage{graphicx,verbatim}
\usepackage{pgf-pie}
\usepackage{array}
\usepackage{float}
\usepackage{amsmath,amssymb}
\usepackage{mathtools}
\usepackage{bm}
\usepackage{overpic}
\usepackage{filecontents}
\usepackage{pgfplotstable}
\usetikzlibrary{positioning}
\usepackage[colorlinks=true, linkcolor=black, citecolor=black, urlcolor=black]{hyperref}

\definecolor{pastelpink}{RGB}{255, 160, 200}   
\definecolor{pastelblue}{RGB}{131, 175, 255}   
\definecolor{pastelorange}{RGB}{255, 204, 153} 
\definecolor{pastelpeach}{RGB}{255, 200, 140} 
\definecolor{pastelpurple}{RGB}{186, 153, 255} 
\definecolor{lightbluegray}{RGB}{180, 195, 210} 
\definecolor{matplotlibpurple}{HTML}{6B2895}

\begin{document}
\title{From Lines to Shapes: Geometric-Constrained Segmentation of X-Ray Collimators via Hough Transform}
\titlerunning{Geometric-Constrained Segmentation of X-Ray Collimators}

\author{Benjamin El-Zein \inst{1,2} \and
Dominik Eckert \inst{1} \and
Andreas Fieselmann \inst{1} \and
Christopher Syben \inst{1} \and
Ludwig Ritschl \inst{1} \and
Steffen Kappler \inst{1} \and
Sebastian Stober \inst{2}}
\authorrunning{B. El-Zein et al.}
%
\institute{Siemens Healthineers, 91301 Forchheim, Germany \and
Artificial Intelligence Lab, Otto-von-Guericke University, Universitätsstr. 2, 39106 Magdeburg, Germany \\
\email{benjamin.el-zein@siemens-healthineers.com}}

    
\maketitle              

\begin{abstract} 

Collimation in X-ray imaging restricts exposure to the region-of-interest (ROI) and minimizes the radiation dose applied to the patient. 
The detection of collimator shadows is an essential image-based pre-processing step in digital radiography posing a challenge when edges get obscured by scattered X-ray radiation.
Regardless, the prior knowledge that collimation forms polygonal-shaped shadows is evident.
For this reason, we introduce a deep learning-based segmentation that is inherently constrained to its geometry.
We achieve this by incorporating a differentiable Hough transform-based network to detect the collimation borders and enhance its capability to extract the information about the ROI center. 
During inference, we combine the information of both tasks to enable the generation of refined, line-constrained segmentation masks.
We demonstrate robust reconstruction of collimated regions achieving median Hausdorff distances of $4.3$--$5.0$mm on diverse test sets of real X-ray images.
While this application involves at most four shadow borders, our method is not fundamentally limited by a specific number of edges.


\end{abstract}
\section{Introduction}

\begin{figure}
    \centering
    \makebox[0.25\textwidth][c]{ 
        \begin{subfigure}{0.2\textwidth}
        \begin{overpic}[height=70pt]{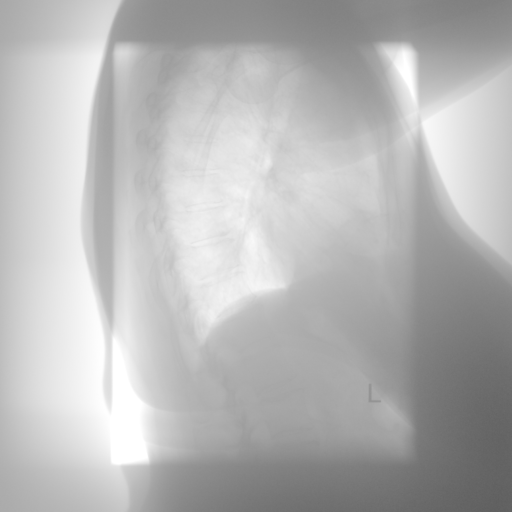}
        \color{matplotlibpurple}
        \linethickness{0.5mm}  
        \multiput(0,45)(5,0){20}{\line(1,0){2}}
        \end{overpic}
        \caption{Image}
        \end{subfigure}
    }
    \hfill
    \makebox[0.25\textwidth][c]{ 
        \begin{subfigure}{0.2\textwidth}
        \begin{overpic}[height=70pt]{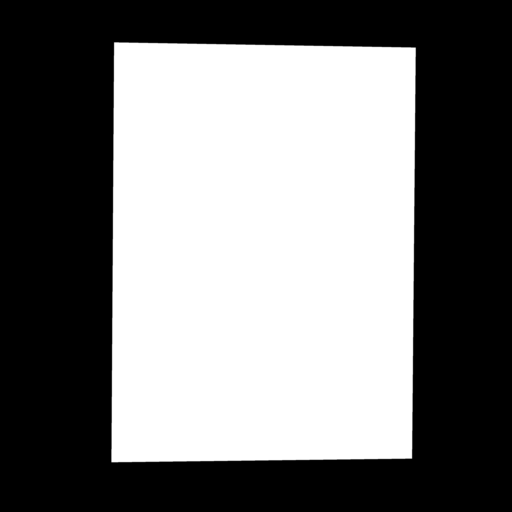}
        \color{blue}
        \linethickness{0.5mm}  
        \multiput(0,45)(5,0){20}{\line(1,0){2}}
        \end{overpic}
        \caption{ROI Mask}
        \end{subfigure}
    }
    \hfill
    \makebox[0.3\textwidth][c]{ 
        \begin{subfigure}{0.275\textwidth}
            \includegraphics[width=\textwidth]{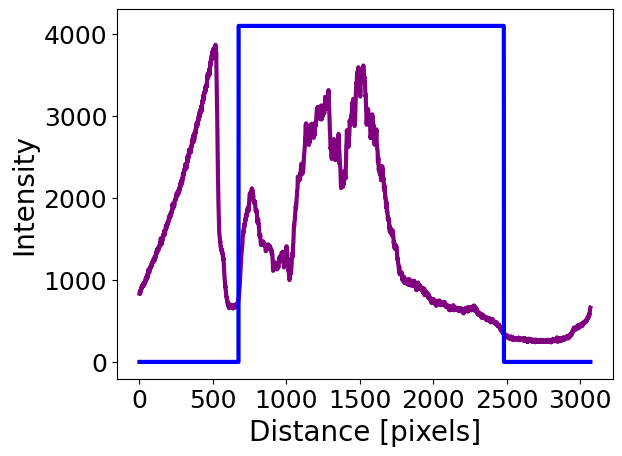}
            \caption{Line Plot}
        \end{subfigure}
    }
    \caption{Example of collimation detection. (a) Example image in log contrast. (b) ROI as binary mask. (c) Line plot of both indicated lines in (a) and (b). Intensities within the collimator shadow exceeding those within the ROI highlight the challenge of the detection task.}
    \label{fig:collimation_example}
\end{figure}

In digital radiography, only relevant parts of the image should be shown to clinicians. Hence, collimator shadows have to be removed from the detector raw image. 
For instance with mobile detectors, there is no information about the detector's position relative to the radiation source \cite{luckner2018estimation}. 
Therefore, an image-based detection mechanism is required for the automated cropping.
Due to physical effects like scattered X-ray radiation or quantum noise, collimator shadows can become hardly distinguishable, as depicted in Fig. \ref{fig:collimation_example}.

Traditional analytical and machine learning methods struggle with robustness, particularly in high-scatter acquisitions \cite{kawashita2003collimation}\cite{mao2014multi}.
Deep learning offers a promising alternative but requires labeled detector raw data, which is scarce.
The simulation pipeline presented in \cite{el2023realistic} addresses this via transfer learning.
By generating synthetic collimated X-ray images, it enables segmentation networks to generalize well to real data mitigating unwanted collimator bias.

Nevertheless, plain convolutional neural networks (CNNs), as in \cite{el2023realistic}, have limitations.
In challenging or out-of-distribution cases, they generate physically inaccurate masks due to lack of geometric awareness.
Collimation projects polygonal ROIs onto the detector plane, resulting in ground truth shapes strictly defined by straight lines.
There are solutions inherently tied to geometry, such as line detection using the differentiable Hough transform \cite{duda1972}.
However, line images leave us with multiple shape options, and hence, the selection of the relevant segment remains unsolved.

The approach in \cite{bachmaier2023robust} presents a Hough transform-based angle-regression for the automated rotation of orthopedic X-ray images. 
Their aim is to detect a single anatomy-specific centerline, which simplifies the selection of line candidates from Hough space to finding the maximum magnitude.
Collimation detection increases the complexity of this mechanism, because its shadows are described by up to four lines.
While Hough transform-based identification of individual lines has been discussed in \cite{zhao2021deep} and \cite{lin2020deep}, their methods tackle RGB-domain problems and do not inherently posses the information for reconstructing the relevant region needed for collimation detection.

In this paper, we propose a novel deep learning architecture that enhances line detection of collimation borders with segmentation of the corresponding region. 
By introducing an additional segmentation branch to the line detection model from \cite{zhao2021deep}, classifying collimated and non-collimated pixels in binary masks, we rewrite its training paradigm.
In our model, Hough space information regularizes the backbone's segmentation.
During inference, we can leverage the aggregated information from both tasks to enable the generation of segmentation masks specifically constrained by the detected lines.
Whereas we demonstrate the incorporation of geometric priors for polygon detection with up to four edges defined by the application to collimation detection, the method is not constrained to a specific number of edges.

\section{Methods}

\subsection{Deep Hough Transform for Collimator Detection}

The Hough transform \cite{duda1972} is an analytical algorithm used in image processing and computer vision for detecting lines within images. 
Its core intuition lies in representing the parametric equations of features, such as lines, in a parameter space rather than Cartesian space, making it more robust even if the features are partially obscured or incomplete.
The fundamental premise of its feature description mechanism is that line features are leveraged, usually by edge filters, such that they can be reliably identified in the Hough space.

\begin{figure}[ht]
    \centering
    \begin{tikzpicture}[
        block/.style={rectangle, anchor=center, draw=pastelblue, fill=white, rounded corners, text centered, minimum height=2em},
        block2/.style={rectangle, anchor=center, draw=matplotlibpurple , fill=white, text centered, minimum height=2em},
        circleblock/.style={circle, anchor=center,draw, text centered, inner sep=0.5pt},
        circleblock2/.style={circle, anchor=center,draw, text centered, inner sep=1pt},
        htlayer/.style={rectangle, anchor=center, draw, fill=teal!30, rounded corners, text centered, minimum height=2em},
        conv/.style={rectangle, anchor=center, draw, text centered, minimum height=1em},
        arrow/.style={->, thick},
        arrow2/.style={<->, thick},
        line/.style={-, thick},
        img/.style={anchor=center, inner sep=0}
    ]
    \node[img] (input) at (0,0) {\includegraphics[width=1.25cm]{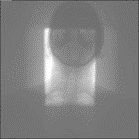}};

    \node[block] (encoder_stage_1) at ($(input.center) + (0, 3)$) {ResBlock};
    \node[block] (encoder_stage_2) at ($(encoder_stage_1.center) + (0, 1.5)$) {ResBlock};
    \node[block] (encoder_stage_3) at ($(encoder_stage_2.center) + (0, 1.5)$) {ResBlock};
    \node[block] (encoder_stage_4) at ($(encoder_stage_3.center) + (0, 1.5)$) {ResBlock};

    \draw[arrow] (input.north) -- (encoder_stage_1.south);
    \draw[arrow] (encoder_stage_1.north) -- (encoder_stage_2.south);
    \draw[arrow] (encoder_stage_2.north) -- (encoder_stage_3.south);
    \draw[arrow] (encoder_stage_3.north) -- (encoder_stage_4.south);

    \node[block] (fpn_stage_4) at ($(encoder_stage_4.center -| encoder_stage_3.center) + (1.75, 0)$) {Stage 1};
    \node[block] (fpn_stage_3) at ($(encoder_stage_3.center -| fpn_stage_4.center) + (0, 0)$) {Stage 2};
    \node[block] (fpn_stage_2) at ($(encoder_stage_2.center -| fpn_stage_3.center) + (0, 0)$) {Stage 3};
    \node[block] (fpn_stage_1) at ($(encoder_stage_1.center -| fpn_stage_2.center) + (0, 0)$) {Stage 4};

    \node[circleblock] (plus3) at ($(fpn_stage_3)!0.5!(fpn_stage_4)$) {+};
    \node[circleblock] (plus2) at ($(fpn_stage_2)!0.5!(fpn_stage_3)$) {+};
    \node[circleblock] (plus1) at ($(fpn_stage_1)!0.5!(fpn_stage_2)$) {+};
    
    \draw[arrow] (fpn_stage_4.south) -- (plus3.north);
    \draw[arrow] ($( encoder_stage_4.center |- encoder_stage_4.center) + (0, -0.75)$) -- (plus3.west);
    \draw[arrow] (plus3.south) -- (fpn_stage_3.north);
    \draw[arrow] (fpn_stage_3.south) -- (plus2.north);
    \draw[arrow] ($( encoder_stage_3.center |- encoder_stage_3.center) + (0, -0.75)$) -- (plus2.west);
    \draw[arrow] (plus2.south) -- (fpn_stage_2.north);
    \draw[arrow] (fpn_stage_2.south) -- (plus1.north);
    \draw[arrow] ($( encoder_stage_2.center |- encoder_stage_2.center) + (0, -0.75)$) -- (plus1.west);
    \draw[arrow] (plus1.south) -- (fpn_stage_1.north);

    \node[block2] (sobel_1) at ($(fpn_stage_1.center -| fpn_stage_1.center) + (3.25, 0)$) {Sobel};
    \node[block2] (sobel_2) at ($(sobel_1.center |- fpn_stage_2.center) + (0, 0)$) {Sobel};
    \node[block2] (sobel_3) at ($(sobel_2.center |- fpn_stage_3.center) + (0, 0)$) {Sobel};
    \node[block2] (sobel_4) at ($(sobel_3.center |- fpn_stage_4.center) + (0, 0)$) {Sobel};
    \draw[arrow] (fpn_stage_1.east) -- (sobel_1.west);
    \draw[arrow] (fpn_stage_2.east) -- (sobel_2.west);
    \draw[arrow] (fpn_stage_3.east) -- (sobel_3.west);
    \draw[arrow] (fpn_stage_4.east) -- (sobel_4.west);

    \node[block2] (ht_1) at ($(sobel_1.center -| sobel_1.center) + (1.1, 0)$) {HT};
    \node[block2] (ht_2) at ($(ht_1.center |- fpn_stage_2.center) + (0, 0)$) {HT};
    \node[block2] (ht_3) at ($(ht_2.center |- fpn_stage_3.center) + (0, 0)$) {HT};
    \node[block2] (ht_4) at ($(ht_3.center |- fpn_stage_4.center) + (0, 0)$) {HT};
    \draw[arrow] (sobel_1.east) -- (ht_1.west);
    \draw[arrow] (sobel_2.east) -- (ht_2.west);
    \draw[arrow] (sobel_3.east) -- (ht_3.west);
    \draw[arrow] (sobel_4.east) -- (ht_4.west);

    \node[block] (ctx_1) at ($(ht_1.center -| ht_1.center) + (1.1, 0)$) {Refine};
    \node[block] (ctx_2) at ($(ctx_1.center |- fpn_stage_2.center) + (0, 0)$) {Refine};
    \node[block] (ctx_3) at ($(ctx_2.center |- fpn_stage_3.center) + (0, 0)$) {Refine};
    \node[block] (ctx_4) at ($(ctx_3.center |- fpn_stage_4.center) + (0, 0)$) {Refine};
    \draw[arrow] (ht_1.east) -- (ctx_1.west);
    \draw[arrow] (ht_2.east) -- (ctx_2.west);
    \draw[arrow] (ht_3.east) -- (ctx_3);
    \draw[arrow] (ht_4.east) -- (ctx_4.west);
    
    \node[fill, circleblock2] (dot1) at ($(fpn_stage_4.east)!0.2!(sobel_4.west)$) {};
    \node[fill, circleblock2] (dot2) at ($(fpn_stage_3.east)!0.4!(sobel_3.west)$) {};
    \node[fill, circleblock2] (dot3) at ($(fpn_stage_2.east)!0.6!(sobel_2.west)$) {};
    \node[fill, circleblock2] (dot4) at ($(fpn_stage_1.east)!0.8!(sobel_1.west)$) {};

    \node[block2] (int1) at ($(plus3.center -| dot1.center) + (0, 0)$) {I};
    \node[block2] (int2) at ($(plus2.center -| dot2.center) + (0, 0)$) {I};
    \node[block2] (int3) at ($(plus1.center -| dot3.center) + (0, 0)$) {I};

    \draw[arrow] (dot1.south) -- (int1.north);
    \draw[arrow] (dot2.south) -- (int2.north);
    \draw[arrow] (dot3.south) -- (int3.north);

    \node[circleblock2] (c) at ($(fpn_stage_1.east)!0.5!(sobel_1.west) + (0, -1)$) {c};
    \draw[line] (int1.south) -- ($( int1.center |- c.center)$);
    \draw[line] (int2.south) -- ($( int2.center |- c.center)$);
    \draw[line] (int3.south) -- ($( int3.center |- c.center)$);
    \draw[line] (dot4.south) -- ($( dot4.center |- c.center)$);
    \draw[arrow] ($( int1.center |- c.center)$) -- (c.west);
    \draw[arrow] ($( dot4.center |- c.center)$) -- (c.east);

    \node[block2] (int4) at ($( c.center |- input.center)$) {I};
    \draw[arrow] (c.south) -- (int4.north);

    \node[block] (onebyone_1) at ($(dot4.center |- int4.center) + (0.1, 0)$) {1x1};
    \draw[arrow] (int4.east) -- (onebyone_1.west);

    \node[img] (mask) at ($( onebyone_1.center -| sobel_1.center) + (0.5, 0)$) {\includegraphics[width=1.25cm]{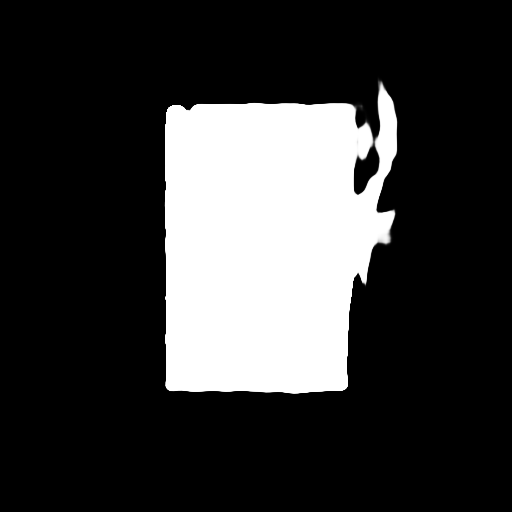}};

    \draw[arrow] (onebyone_1.east) -- (mask.west);


    \node[circleblock2] (c2) at ($( plus2.center -| ctx_4.center) + (0.75, 0)$) {c};
    \draw[line] (ctx_4.east) -- ($( c2.center |- ctx_4.center)$);
    \draw[line] (ctx_3.east) -- ($( c2.center |- ctx_3.center)$);
    \draw[line] (ctx_2.east) -- ($( c2.center |- ctx_2.center)$);
    \draw[line] (ctx_1.east) -- ($( c2.center |- ctx_1.center)$);
    \draw[arrow] ($( c2.center |- ctx_1.center)$) -- (c2.south);
    \draw[arrow] ($( c2.center |- ctx_4.center)$) -- (c2.north);
    
    \node[block] (onebyone_2) at ($( c2.center |- c2.center) + (0.75, 0)$) {1x1};
    \draw[arrow] (c2.east) -- (onebyone_2.west);

    \node[block2] (gaussian) at ($(onebyone_2.center |- onebyone_2.center) + (1.25, 0)$) {Gauss};

    \node[img] (hough_output) at ($( gaussian.center |- gaussian.center) + (0, -1.5)$) {\includegraphics[width=1.25cm]{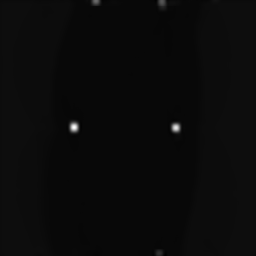}};
    \draw[arrow] (onebyone_2.east) -- (gaussian.west);

    \node[img] (mask_label) at ($( onebyone_2.center |- mask.center)+ (0, 0)$) {\includegraphics[width=1.25cm]{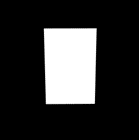}};

    \node[img] (hough_label) at ($( hough_output.center |- mask_label.north) + (0, 0.625)$) {\includegraphics[width=1.25cm]{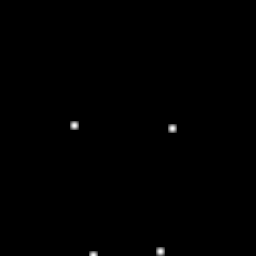}};

    \draw[arrow] (gaussian.south) -- (hough_output.north);
    
    \draw[arrow2] (hough_output.south) -- (hough_label.north) node[midway, left] {Hough Loss};
    \draw[arrow2] (mask.east) -- (mask_label.west) node[midway, above] {Mask Loss};

    \draw[arrow] ($( hough_label.center |- mask_label.center)$) -- (mask_label.east) node[pos=0, below] {Labels};
    \draw[arrow] ($( hough_label.center |- mask_label.center)$) -- (hough_label.south);

\end{tikzpicture}
\caption{Block diagram of our proposed architecture. Purple blocks correspond to known-operators: Sobel filters, I-blocks for interpolations steps, HT incorporate Hough transform layers, and Gaussian smoothing ensuring gradient flow for distinct peak predictions. Blue blocks introduce learnable modules: ResBlocks are the encoder blocks, the stages correspond to those in the FPN, 1x1-blocks are 1x1 convolutions and Refine-blocks are two convolutional layers each for refinement of Hough space artifacts.}
\label{fig:architecture}
\end{figure}

Since in the deep Hough transform (DHT) \cite{zhao2021deep} edge filters are not part of the intrinsic architectures, the convolutional layers of its feature pyramid network (FPN) \cite{lin2017feature} backbone have to learn extracting edge features from the input itself. 
Applied to polygon segmentation, the information about the corresponding region is lost. 
To retain this information we introduce a new branch to the network that leads to a second loss function enforcing segmentation of the ROI on each resolution stage.
As visualized in Fig. \ref{fig:architecture}, the feature maps at different stages are interpolated to match the resolution of the last stage.
Hence, they can be stacked into a single tensor and aggregated into a single-channel output by a 1x1-convolution.
Finally, another interpolation is applied to match the input image resolution, which reduces computational effort compared to directly matching this resolution in the previous step.

At this point, the introduced branch imposes two mutually exclusive properties, leading to an inherent contradiction. Whereas the Hough transform expects an edge image, we enforce its input layers to generate masks.
Motivated by the principle that deep networks converge faster if known operators \cite{maier2019learning} are an intrinsic part of the architecture, we use differentiable Sobel filters to resolve this inconsistency, instead of increasing the networks complexity to learn edge filters.

\begin{figure}[ht]
    \centering
    \begin{tikzpicture}[
        block/.style={rectangle, anchor=center, draw=red, fill=white, text centered, minimum height=2em},
        block2/.style={rectangle, anchor=center, draw=matplotlibpurple , fill=white, text centered, minimum height=2em},
        circleblock/.style={circle, anchor=center,draw, text centered, inner sep=0.5pt},
        circleblock2/.style={circle, anchor=center,draw, text centered, inner sep=1pt},
        htlayer/.style={rectangle, anchor=center, draw, fill=teal!30, rounded corners, text centered, minimum height=2em},
        conv/.style={rectangle, anchor=center, draw, text centered, minimum height=1em},
        arrow/.style={->, thick},
        arrow2/.style={<->, thick},
        line/.style={-, thick},
        img/.style={anchor=center, inner sep=0}
    ]
    \node[img] (input) at (0,0) {\includegraphics[width=1.25cm]{images/fig_2/architecture_dhtregoutput.png}};
    \node[left=5pt of input] {Region Mask};
    
    \node[img] (input2) at ($( input.south -| input.south) + (0, -1.5)$) {\includegraphics[width=1.25cm]{images/fig_2/architecture_hough_output.png}};
    \node[left=5pt of input2] {Hough Space};

        \node[block] (com) at ($( input.east -| input.east) + (1.5, 0)$) {(x, y)};
    \node[above=5pt of com] {Center of Mass};
    \draw[arrow]  (input.east) -- (com.west);

    \node[block2] (iht) at ($( com.center |- input2.center) + (0, 0)$) {IHT};
    \node[below=5pt of iht] {Line-Encoding};
    
    \draw[arrow]  (input.east) -- (com.west);
    \draw[arrow]  (input2.east) -- (iht.west);

    \node[circleblock] (dot) at ($(com.center)!0.5!(iht.center) + (0.85, 0)$) {};

    \draw[line] (com.east) -- ($( dot.center |- com.center)$);
    \draw[line] (iht.east) -- ($( dot.center |- iht.center)$);
    \draw[line] ($( dot.center |- com.center)$) -- ($( dot.center |- iht.center)$);

    \node[img] (input3) at ($(dot.center) + (1.5, 0)$) {
    \begin{overpic}[height=50pt]{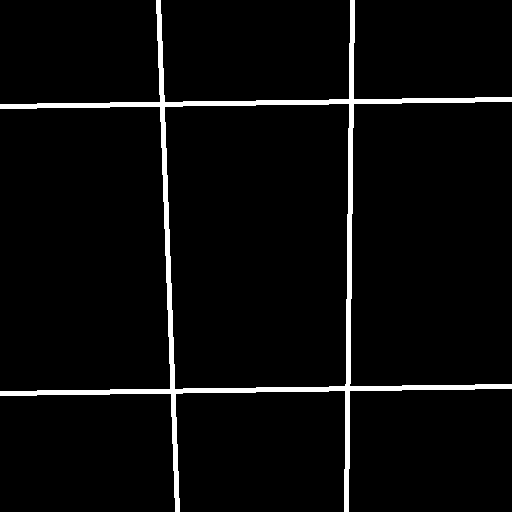}
    \color{red}
    \put(57,57){\circle*{5}}  
    \end{overpic}
    };
    \node[above=5pt of input3] {Fusion};

    \draw[arrow]  (dot.east) -- (input3.west);

    \node[block] (fl) at ($( input3.east -| input3.east) + (1.15, 0)$) {FL};
    \node[above=5pt of fl] {Flood-Fill};

    \node[img] (input4) at ($( fl.east -| fl.east) + (1.4, 0)$) {\includegraphics[width=1.25cm]{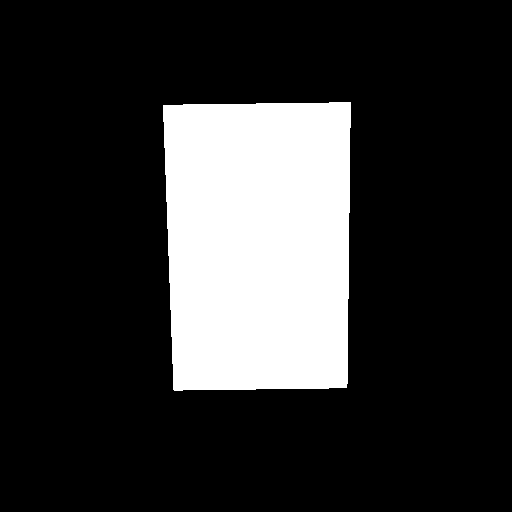}};
    \node[above=5pt of input4] {Prediction};

    \draw[arrow]  (input3.east) -- (fl.west);
    \draw[arrow]  (fl.east) -- (input4.west);

\end{tikzpicture}
\caption{Post-processing for obtaining segmentation masks from our model during inference, utilizing an inverse Hough transform (IHT) as well as the center of mass (x, y) of the region mask. The information provided by both branches is fused to reconstruct shape-constrained predictions via flood-filling.}
\label{fig:rht}
\end{figure}

To obtain the segmentation masks, we adapt the post-processing in \cite{zhao2021deep}, delivering an approach for an inverse Hough transform approach.
After applying a threshold in Hough space, the remaining regions are determined as relevant region blob's.
The selection problem is solved by considering the blobs center of mass as the underlying line.
This yields an image space with lines that form different geometric shapes.
The center of mass of our segmentation branch's mask delivers a robust seed point for the corresponding ROI region.
Therefore, a traditional flood fill algorithm \cite{van2014scikit} can be applied to finally turn the line image into a line-constrained segmentation mask.
A scheme of our adapted post-processing is demonstrated in Fig. \ref{fig:rht}.

\subsection{Collimator Data and Hough Space Annotations}

The data set used for this research consists of 1680 clinical X-ray images.
Initially, the data is divided into the following splits: 1500 images for training, 50 images for validation and 130 images for testing. 
The relatively small proportion of test and validation samples is due to careful selection by hand.
The validation set and a first of three subsets (80 images) for testing are chosen to ensure that all image types shown during training are represented.
The remaining subsets first contain high-absorbing, often line shaped implants like screws (30 images), and second posses detector line-artifacts that may interfere with line detection (20 images).


The augmentation strategy proposed in \cite{el2023realistic} demonstrates that using simulated collimator data is beneficial regarding generalization on real data.
Real collimator properties, including edge blurring, scattered radiation and poisson noise, are simulated onto images that do not contain such shadows, given a randomly generated mask label.
Following this approach, we remove real collimator shadows from our training set, and utilize the simulation pipeline as on-the-fly augmentation for training only.
Thereby, segmentation masks as well as Hough space ground truth are inherently available, and we obtain increased diversity of distinct collimations.
During inference, we apply the model to the real collimations.
We manually annotated the test set to get the required Hough space labels, enabling complete evaluation on real data.

\subsection{Experiments}

\subsubsection{Line Detection for Collimator Edges:} To test effectiveness of the Hough transform-based output in the baseline model and ours for classifying correct collimator edges, we follow the strategy in \cite{zhao2021deep}.

The process of assigning true pairs from the set of detected lines to the set of ground truth lines is formulated as a maximum bipartite graph matching problem \cite{kuhn1955hungarian}. 
This approach interprets the relationships between elements of the two sets as a bipartite graph and finds the largest possible pairing, ensuring that each element is matched at most once. 
Decisions to accept or reject potential pairs rely on a line metric and a threshold.
\cite{zhao2021deep} proposes the EA (Euclidean and Angular) score measuring line similarity, which allows for a threshold sweep to estimate the F1 score, precision, and recall.
Hence, we can evaluate the ability of networks to distinguish relevant collimator edges.
The baseline model as well as ours are initialized with random parameters, trained using z-normalization and optimized by ADAM.
After empirical estimation, we define the learning rate to be $1 \times 10^{-5}$, while the batch size used is set to be 20.
We define our total loss function to be the addition of the Dice similarity coefficient (Mask Loss) as segmentation loss, and the multi-scale structural similarity index (Hough Loss) as loss within Hough space:

\begin{equation}
  L_T(\Vec{X}, \Vec{Y}) = \delta \cdot Mask Loss(\Vec{X}, \Vec{Y}) + \epsilon \cdot Hough Loss(\Vec{X}, \Vec{Y})
  \label{equ:lossfunc}
\end{equation}

where $\Vec{X}$ and $\Vec{Y}$ are prediction and ground truth, as well as $\delta$ and $\epsilon$ being hyperparameters of the total loss function $L_T$, both being empirically set to 1.

Note that in \cite{zhao2021deep}, the model's Hough space threshold $t$ is set to zero, and variations are not considered.
We aim for leveraging this.
Therefore, we calibrate both networks based on the F1 score using the synthetic validation data set from training.
For consistency, the original method is also demonstrated and referred to as the baseline, whereas the calibrated model is referred to as DHT.

\subsubsection{From Lines to Shapes:} To analyze the transfer from lines to shapes, we evaluate the performance on the segmentation masks first in terms of Dice score. 
The deep learning approach in \cite{el2023realistic} serves as state-of-the-art collimator detection reference.
In addition to that, we would like to investigate the boundary error of our line-constrained network.
Therefore, we estimate the average Hausdorff distance in millimeter-scale within the detector plane.
This allows for analysis on the impact of geometric constraints on clinical use.

\section{Results}

\subsection{Collimator Edge Classification}

First, the goal is to obtain insight into the effectiveness of interpreting collimator segmentation as an edge detection problem, since it incorporates the prior knowledge about collimation shapes.
We compare the line detection baseline to our adaptation of it (DHT), both trained on synthetic collimator data.
Table \ref{table:Classification} shows the average F1 score, precision and recall of every test set containing real collimators.
It can be concluded that calibrating on validation data significantly improves the performance of the baseline model.
Although recall decreases for higher thresholds, the significant increase in precision suggests that the high recall previously resulted from predicting more lines than necessary to ensure the relevant edges were included in the detected set.
Thus, a higher threshold refines the Hough space encoding to better focus relevant lines.
Our model delivers competitive performance, and increases F1 score in two of three cases.
However, it seems that all of the models have difficulties with the artifacts test.

\begin{table}[h]
    \centering
    \caption{Average F1 score, precision, and recall on collimator edge classification for the considered models, given the threshold $t$ used during post-processing.}
    \begin{adjustbox}{width=\textwidth}
    \begin{tabular}{@{} lccccccccc @{}}
        \toprule
        \multirow{2}{*}{\textbf{Model [$t$]} }
        & \multicolumn{3}{c}{\textbf{General}} 
        & \multicolumn{3}{c}{\textbf{Artifacts}} 
        & \multicolumn{3}{c}{\textbf{Implants}} \\
        \cmidrule(lr){2-4} \cmidrule(lr){5-7} \cmidrule(lr){8-10}
        & F1 & Precision & Recall 
        & F1 & Precision & Recall 
        & F1 & Precision & Recall \\
        \midrule
        Baseline [$t{=}0$] 
            & $0.7803$ & $0.6769$ & $0.9210$ 
            & $0.6070$ & $0.5974$ & $0.6170$ 
            & $0.7623$ & $0.6526$ & $0.9164$ \\
        
        DHT [$t{=}0.35$] 
            & $0.8605$ & $0.8788$ & $0.8429$ 
            & $0.6563$ & $0.7633$ & $0.5756$ 
            & $\mathbf{0.8718}$ & $0.8627$ & $0.8810$ \\
        \rowcolor{gray!10}
        Ours [$t{=}0.2$] 
            & $\mathbf{0.8875}$ & $0.8640$ & $0.9122$ 
            & $\mathbf{0.7155}$ & $0.8221$ & $0.6334$ 
            & $0.8592$ & $0.8503$ & $0.8684$ \\
        \bottomrule
    \end{tabular}
    \end{adjustbox}
    \label{table:Classification}
\end{table}

\subsection{Collimator Segmentation}

We first compare quantitative results of our line-constrained model to the plain segmentation approach (SegNet) for collimator segmentation from \cite{el2023realistic} by using the same data set.
In Table \ref{table:DiceScores}, the Dice scores of both models for the three test sets are demonstrated.
Although both networks exhibit comparable performance in quantitative evaluation, our approach offers a substantial qualitative advantage from an application perspective. 
This improvement arises from the geometric shape constraints that are absent in SegNet, ensuring more reliable, interpretable, and application-specific results.

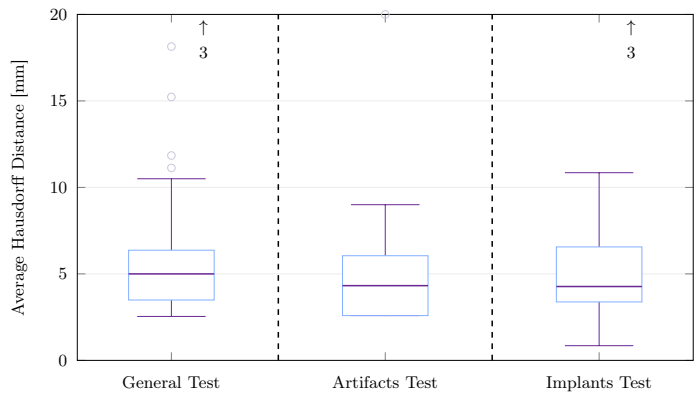
\begin{figure}[htbp]
\centering
\begin{adjustbox}{width=0.75\textwidth}
\begin{tikzpicture}
    \centering
    \definecolor{network1}{RGB}{173,216,230}
    \definecolor{network2}{RGB}{135,206,235}
    \definecolor{network3}{RGB}{30,144,255}
    \definecolor{network4}{RGB}{65,105,225}

    \definecolor{network1}{RGB}{200,200,220}
    \definecolor{network2}{RGB}{180,180,210}
    \definecolor{network3}{RGB}{100,100,160}
    \definecolor{network4}{RGB}{50,50,110}

    \begin{axis}[
        boxplot/draw direction=y,
        xtick={1, 3, 5},
        xticklabels={General Test, Artifacts Test, Implants Test},
        x tick label style={yshift=-5},
        ylabel={Average Hausdorff Distance [mm]},
        y label style={yshift=5},
        y tick label style={xshift=-2},
        width=13cm,
        height=8cm,
        ymin=0,
        ymax=20,
        ymajorgrids,
        grid style={opacity=0.3},
        legend style={at={(1.015,1)},|-|,anchor=north west},
        legend image post style={sharp plot, mark=square*},
        boxplot/every median/.append style={matplotlibpurple, thick, solid},
        boxplot/every box/.style={draw=pastelblue, fill=white, solid},
        boxplot/every whisker/.style={solid, matplotlibpurple},
        boxplot/every whisker/.append style={solid}
    ]
    
    \addplot[
        boxplot prepared={
            lower whisker= 2.54,
            lower quartile= 3.49,
            median=5.00,
            upper quartile= 6.37,
            upper whisker= 10.5
        },
        fill=network1,
        draw=network1,
        mark=o,
        boxplot/draw position=1
    ] coordinates {(1, 11.127392686519155)(1, 11.844)(1, 18.144)(1, 15.227999999999998)};
     
    \addplot[dashed, thick] coordinates {(2,0) (2,20)};

    \addplot[
        boxplot prepared={
            lower whisker= 2.59,
            lower quartile= 2.59,
            median=4.32,
            upper quartile= 6.05,
            upper whisker= 9
        },
        fill=network1,
        draw=network1,
        mark=o,
        boxplot/draw position=3
    ] coordinates{(3, 20)};
    
    \addplot[dashed, thick] coordinates {(4,0) (4,20)};

    \addplot[
        boxplot prepared={
            lower whisker= 0.85,
            lower quartile= 3.38,
            median=4.27,
            upper quartile= 6.56,
            upper whisker= 10.85
        },
        fill=network1,
        draw=network1,
        mark=o,
        boxplot/draw position=5
    ] coordinates {};

    \node at (axis cs:1.3,20) [below] {$\uparrow$};
    \node at (axis cs:1.3,18.5) [below] {$3$};
    \node at (axis cs:5.3,20) [below] {$\uparrow$};
    \node at (axis cs:5.3,18.5) [below] {$3$};


    \end{axis}
\end{tikzpicture}
\end{adjustbox}
\caption{Average Hausdorff distance box plot of our model applied to the real collimator test sets.}
  \label{fig:ResultsBoxPlot}
\end{figure}

\begin{figure}[htbp]
    \centering
    \begin{table}[H]
        \centering
        \begin{adjustbox}{width=0.78\textwidth}
        \begin{tabular}{>{\centering\arraybackslash}m{0.15\textwidth} >{\centering\arraybackslash}m{0.15\textwidth} >{\centering\arraybackslash}m{0.15\textwidth} >{\centering\arraybackslash}m{0.15\textwidth} >{\centering\arraybackslash}m{0.15\textwidth}}
        \toprule
        Example Image & Label & SegNet & DHT & Ours \\  
        \midrule
        \centering \includegraphics[width=0.15\textwidth]{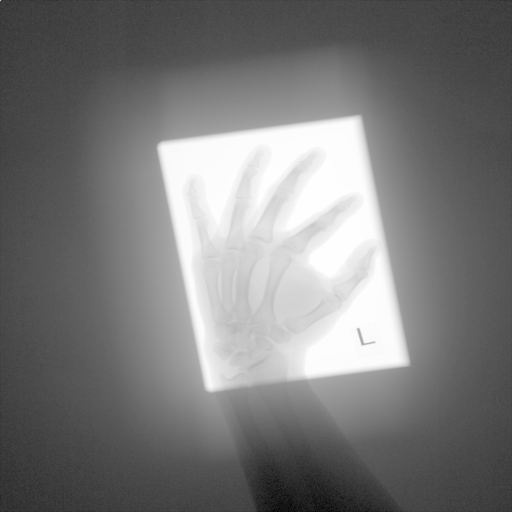}  &
       \centering \includegraphics[width=0.15\textwidth]{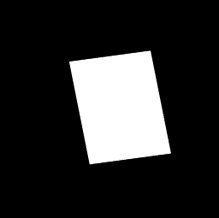} &
       \centering \includegraphics[width=0.15\textwidth]{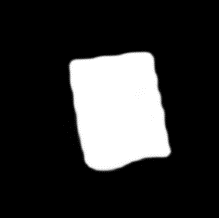} &
       \centering \includegraphics[width=0.15\textwidth]{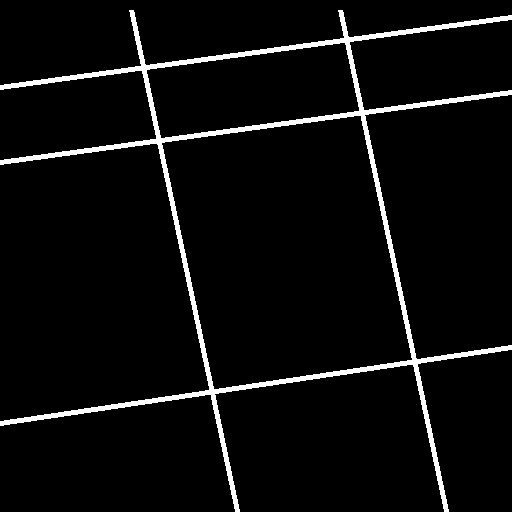} & 
       \includegraphics[width=0.15\textwidth]{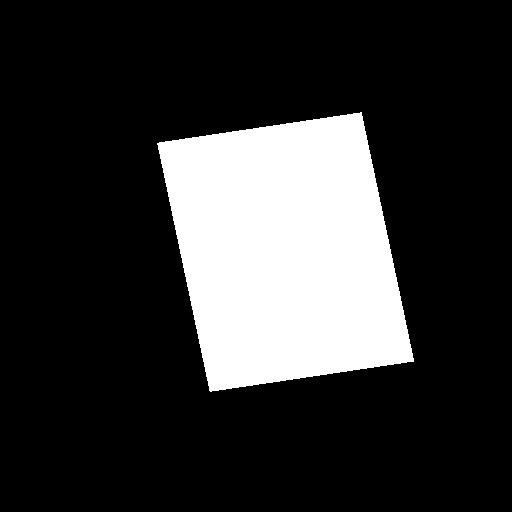} 
        \\
        \midrule
        \centering \includegraphics[width=0.15\textwidth]{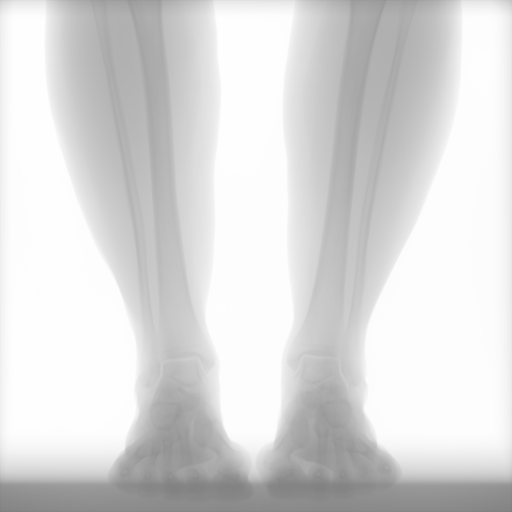}  &
       \centering \includegraphics[width=0.15\textwidth]{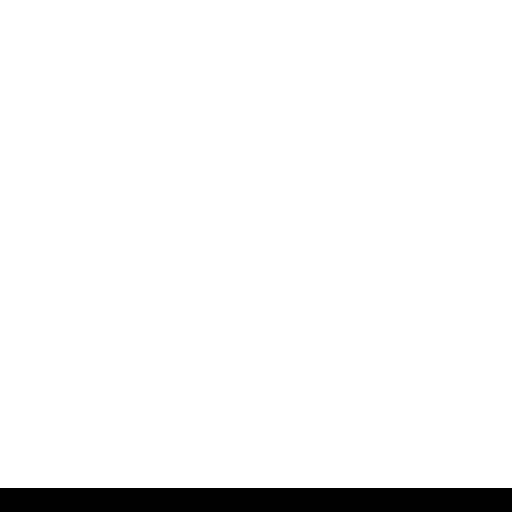} &
       \centering \includegraphics[width=0.15\textwidth]{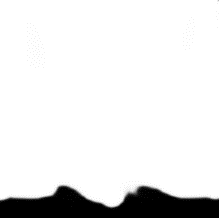} &
       \centering \includegraphics[width=0.15\textwidth]{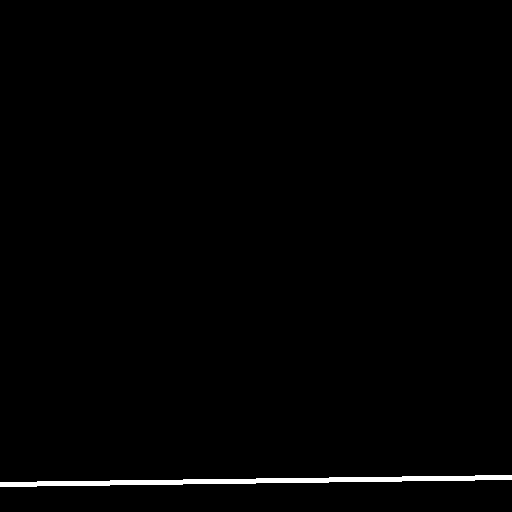} & 
       \includegraphics[width=0.15\textwidth]{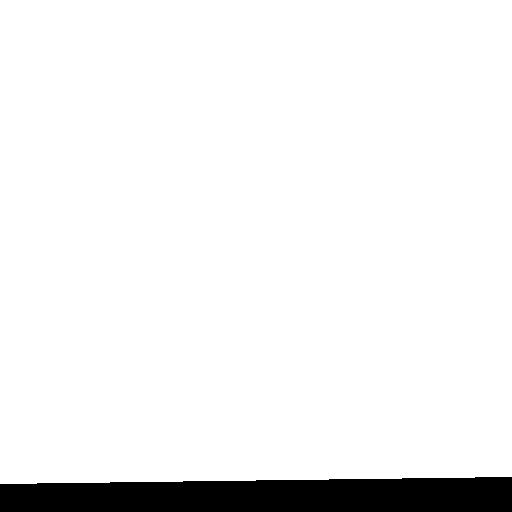} 
        \\
        \bottomrule
        \end{tabular}
    \end{adjustbox}
    \end{table}
    \caption{Qualitative prediction comparison of SegNet, DHT and ours for two example images. Whereas SegNet lacks in geometric awareness at the border regions, DHT delivers line images that leave us with multiple shape options. Our model combines the segmentation aspect of SegNet with the shape information of DHT and predicts line-constrained polygons.}
    \label{fig:qual_example}
\end{figure}

Furthermore, we focus on the boundary error. 
Since this is only methodologically meaningful for our line-constrained network, we report its performance using the average Hausdorff distance. 
To improve result interpretability, we estimate its scores in millimeter-scale on the detector plane.
A corresponding box plot is presented in Fig. \ref{fig:ResultsBoxPlot}.
The median Hausdorff distance for all test sets is $5$mm or lower, indicating solid performance.
To qualitatively reveal the key improvement of our approach the predictions of both reference models (SegNet, DHT) are compared to ours in Fig. \ref{fig:qual_example}.
Two example images are shown with the corresponding ground truth.
Whereas SegNet suffers from unconstrained boundaries which get mislead by high uncertainty due to intensity variations, DHT is able to predict lines that amongst other shapes show the required ROI.
Our network is able to overcome this limitation and predicts the whole relevant and shape-constrained region.

\begin{table}[htbp]
    \centering
    \caption{Dice score performance (mean $\pm$ standard deviation)  of SegNet and our model on real test sets.}
    \begin{adjustbox}{width=0.65\textwidth}
    \begin{tabular}{lccc}
        \toprule
        \textbf{Model} & 
        \textbf{General} & 
        \textbf{Artifacts} & 
        \textbf{Implants} \\
        \midrule
        SegNet & 
        $0.9718 \pm 0.027$ & 
        $0.9778 \pm 0.025$ & 
        $0.9494 \pm 0.071$ \\
        \rowcolor{gray!10}
        Ours & 
        $\mathbf{0.9763} \pm 0.020$ & 
        $\mathbf{0.9910} \pm 0.007$ & 
        $\mathbf{0.9685} \pm 0.077$ \\
        \bottomrule
    \end{tabular}
    \end{adjustbox}
    \label{table:DiceScores}
\end{table}

\section{Discussion}


The presented approach demonstrates significant advancements in collimation edge detection and segmentation.
To our knowledge, this is the first deep learning-based method for collimator segmentation that incorporates constraints based on actual geometry.
Our strategy not only proves to be effective when compared to a dedicated line detection model on X-ray data, but, more importantly, leads to improved detection quality compared to a leading collimator segmentation method.
We evaluated performance on a comprehensive real-world data set, including two more challenging sub sets: one with line-shaped implants and another with detector line-artifacts.
Our model achieved the best results on the artifacts set, demonstrating robust detection of relevant lines despite the presence of other interfering lines.
This may be attributed to lower amount of scatter in these cases, allowing the model to focus more effectively on relevant line structures.
Future work will focus on developing a fully end-to-end solution to reduce the influence of human-defined heuristics and potential biases introduced during post-processing, while ensuring tighter integration across system components.
Finally, although shape generation has been achieved, introducing stricter constraints is needed to reduce degrees of freedom and enable more precise, controllable outputs.

\noindent\textbf{Disclaimer:} The concepts and information presented in this paper are based on research and are not commercially available.

\newpage
%
%
%
%





\bibliographystyle{splncs04}
\bibliography{literature}

\end{document}